\documentclass[10pt,twocolumn,letterpaper]{article}

\usepackage{iccv}
\usepackage{times}
\usepackage{epsfig}
\usepackage{graphicx}
\usepackage{amsmath}
\usepackage{amssymb}
\usepackage{multirow}
\usepackage{tabularx}
\usepackage{comment}
\usepackage{siunitx}
\usepackage{capt-of}
\usepackage{wrapfig}
\usepackage{pdfpages}
\usepackage[dvipsnames]{xcolor}

\usepackage[breaklinks=true,bookmarks=false]{hyperref}

\iccvfinalcopy 

\def\iccvPaperID{1228} 

\newcolumntype{H}[1]{>{\hsize=#1\hsize\arraybackslash}X}

\newcommand{\mybullet}{\vspace{0.05cm}\noindent $\bullet$\ }
\newcommand{\mybulletend}{\vspace{0.05cm}}

\newcommand{\sachini}[1]{\textcolor{magenta}{[sachini: {#1}]}}
\newcommand{\yasu}[1]{\textcolor{orange}{[yasu: {#1}]}}

\newcommand{\mysubsubsection}[1]{\vspace{0.1cm} \noindent {\bf #1}:}
\newcommand{\mysubsubsubsection}[1]{\vspace{0.1cm} \noindent {\bf #1}:}

\definecolor{darkgreen)}{rgb}{0.0, 0.5, 0.0}

\begin{document}

\title{RoNIN: Robust Neural Inertial Navigation in the Wild:\\Benchmark, Evaluations, and New Methods}

\author{Hang Yan\footnotemark[1]\\
Washington University in St. Louis\\
St. Louis, USA\\
{\tt\small yanhang@wustl.edu}
\and
Sachini Herath \footnotemark[1]\\
Simon Fraser University\\
BC, Canada\\
{\tt\small sherath@sfu.ca}
\and
Yasutaka Furukawa\\
Simon Fraser University\\
BC, Canada\\
{\tt\small furukawa@sfu.ca}
}

\twocolumn[{%
\renewcommand\twocolumn[1][]{#1}%
\maketitle

\begin{center}
    \centering
    \includegraphics[width=\textwidth]{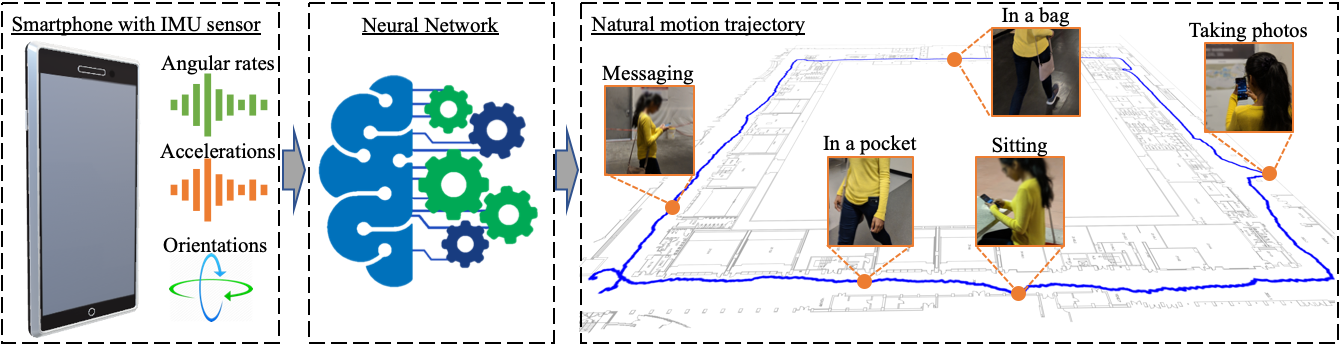}
 \captionof{figure}{Inertial navigation is the problem of estimating the position and orientation of a moving subject only from a sequence of IMU sensor data. This paper presents a new benchmark, new algorithms, and extensive evaluations of existing techniques for inertial navigation.
 }
 \label{fig:teaser}
\end{center}%
}]
\renewcommand{\thefootnote}{\fnsymbol{footnote}}
\footnotetext[1]{Indicates equal contributions.}

\renewcommand*{\thefootnote}{\arabic{footnote}}
\begin{abstract}
This paper sets a new foundation for data-driven inertial navigation research, where the task is the estimation of positions and orientations of a moving subject from a sequence of IMU sensor measurements.
%
More concretely, the paper presents 1) a new benchmark containing more than 40 hours of IMU sensor data from 100 human subjects with ground-truth 3D trajectories under natural human motions; 2) novel neural inertial navigation architectures, making significant improvements for challenging motion cases;
and 3) qualitative and quantitative evaluations of the competing methods over three inertial navigation benchmarks.
%
We will share the code and data to promote further research.\footnote{Project website: http://ronin.cs.sfu.ca/}
\end{abstract}

\section{Introduction}

An inertial measurement unit (IMU), often a combination of accelerometers, gyroscopes, and magnetometers, plays an important role in a wide range of navigation applications. In Virtual Reality, IMU sensor fusion produces real-time orientations of head-mounted displays. In Augmented Reality applications (e.g., Apple ARKit~\cite{arkit}, Google ARCore~\cite{arcore}, or Microsoft HoloLens\cite{hololens}), IMU augments SLAM~\cite{mur2017visualslam, leutenegger2015keyframe, forster2017manifold} by resolving scale ambiguities and providing motion cues in the absence of visual features.
%
UAVs, autonomous cars, humanoid robots, and smart vacuum cleaners are other emerging domains, utilizing IMUs for enhanced navigation, control, and beyond.

Inertial navigation is the ultimate form of IMU-based navigation, whose task is to estimate positions and orientations of a moving subject only from a sequence of IMU sensor measurements (See Fig. \ref{fig:teaser}). Inertial navigation has been a dream technology for academic researchers and industrial engineers, as IMUs 1) are energy-efficient, capable of running 24 hours a day; 2) work anywhere even inside pockets; and 3) are in every smartphone, which everyone carries everyday all the time.



Most existing inertial navigation algorithms require unrealistic constraints that are incompatible with everyday smartphone usage scenarios.
For example, an IMU must be attached to a foot to enable the {\it zero speed update} heuristic (i.e., a device speed becomes 0 every time a foot touches the ground)~\cite{jimenez2009zupt}.
Step counting methods assume that the IMU is rigidly attached to a body and a subject must walk forward so that the motion direction becomes a constant in device coordinate frame~\cite{step_counting}.

Data-driven approaches have recently made a breakthrough in loosing these constraints~\cite{yan2018ridi, chen2018oxiod}
where the acquisition of IMU sensor data and ground-truth motion trajectories allows supervised learning of direct motion parameters (e.g., a velocity vector from IMU sensor history).
%
This paper seeks to take data-driven inertial navigation research to the next level via the following three contributions.

\vspace{0.05cm}
\mybullet
The paper provides the largest inertial navigation database 
consisting of more than 42.7 hours of IMU and ground-truth 3D motion data from 100 human subjects. Our data acquisition protocol allows users to handle smartphones naturally as in real day-to-day activities.

\mybullet
The paper presents novel neural architectures for inertial navigation, making significant improvements for challenging motion cases over the existing best method.

\mybullet
The paper presents extensive qualitative and quantitative evaluations of existing baselines and state-of-the-art methods on the three benchmarks. 

\vspace{0.1cm}
We will share the code and data to promote further research in a hope to establish an ultimate anytime anywhere navigation system for everyone's smartphone.
\section{Related Work}

We group inertial navigation algorithms into three categories based on their use of priors.

\mysubsubsubsection{Physics-based (no priors)} 
IMU double integration is a simple idea for inertial navigation.
Given the device orientation (e.g., via Kalman filter\cite{kalman1960new} on IMU signals), one subtracts the gravity from the device acceleration, integrates the residual accelerations once to get velocities, and integrates once more to get positions. Unfortunately, sensor biases explode quickly in the double integration process, and these systems do not work in practice without additional constraints. A foot mounted IMU with zero speed update is probably the most successful example, where the sensor bias can be corrected subject to a constraint that the velocity must become zero whenever a foot touches the ground.

\mysubsubsubsection{Heuristic priors} Human motions are highly repetitive. Most existing inertial navigation research seeks to find heuristics exploiting such motion regularities. Step counting is a popular approach assuming that 1) An IMU is rigidly attached to a body; 2) The motion direction is fixed with respect to the IMU; and 3) The distance of travel is proportional to the number of foot-steps. The method produces impressive results in a controlled environment where these assumptions are assured. More sophisticated approaches utilize principal component analysis~\cite{janardhanan2014attitude} or frequency domain analysis~\cite{kourogi2014method} to infer motion directions. However, these heuristic based approaches do not match up with the robustness of emerging data-driven methods~\cite{yan2018ridi}.
%

\mysubsubsubsection{Data-driven priors}
Robust IMU double integration (RIDI) was the first data driven Inertial navigation method~\cite{yan2018ridi}. RIDI focuses on regressing velocity vectors in a device coordinate frame, while relying on traditional sensor fusion methods to estimate device orientations. 
%
RIDI works for complex motion cases such as backward-walking, significantly expanding the operating ranges of the inertial navigation system.
IONet is a neural network based approach, which regresses the velocity magnitude and the rate of motion-heading change
without relying on external device orientation information~\cite{chen2018ionet}.


\mysubsubsubsection{Inertial navigation datasets}
RIDI dataset utilized a phone with 3D tracking capability (Lenovo Phab Pro 2) to collect IMU-motion data under four different phone placements (i.e., a hand, a bag, a leg pocket, and a body). The Visual Inertial SLAM produced the ground-truth motion data. The data was collected by 10 human subjects, totalling 2.5 hours.
IONet dataset, namely OXIOD used
a high precision motion capture system (Vicon) under four different phone placements (i.e., a hand, a bag, a pocket, and a trolley)~\cite{chen2018oxiod}. The data was collected by five human subjects, totalling 14.7 hours. 

The common issue in these datasets is the reliance on a single device for both IMU data and the ground-truth motion acquisition. 
The phone must have a clean line-of-sight for Visual Inertial SLAM or must be clearly visible for the Vicon system all the time, prohibiting natural phone handling especially for a bag and a leg pocket scenarios.
This paper presents a new IMU-motion data acquisition protocol that utilizes two smartphones to overcome these issues.

\section{The RoNIN dataset}

\begin{figure*}[tb]
    \centering
    \includegraphics[width=\linewidth]{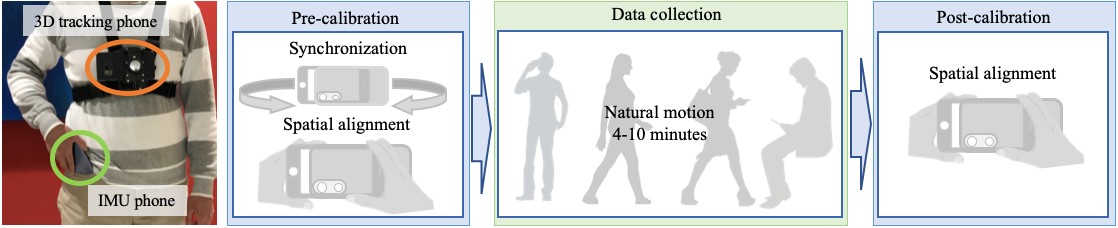}
    \caption{Left: Our two-device protocol allows subjects to handle the IMU phone freely, while relying on the body mounted 3D tracking phone to produce ground-truth motion trajectories. Right: Before the data-collection, we calibrate IMU sensor biases and align two devices in space and time. After the data collection, we spatially align two devices again to assess the accumulation errors in the IMU device orientation.
    See the supplementary document for our full data collection steps.}
    \label{fig:data_collection}
\end{figure*}



Scale, diversity and fidelity are the three key factors in building a next-generation inertial navigation database.
In comparison to the current largest database OXIOD~\cite{chen2018oxiod}, our dataset boasts

\mybullet Scale:
42.7 hours of IMU-motion data (2.9 times more than OXIOD) over 276 sequences in 3 buildings,

\mybullet Diversity:
100 human subjects (20 times more than OXIOD)  with three Android devices,\footnote{Asus Zenfone AR, Samsung Galaxy S9 and Google Pixel 2 XL, where the first device uses a ICM20602 IMU sensor from InvenSense and the latter two use the same LSM6DSL sensor from STMicro.}

\mybullet Fidelity: subjects handle devices naturally as in real day-to-day activities such as carrying inside a bag, placing deep inside a pocket, or picking up by hand, while walking, sitting or wandering around.

\mybulletend

We have developed a two-device data acquisition protocol, where we use a harness to attach a 3D tracking phone (Asus Zenfone AR) to a body and let subjects handle the other phone freely for IMU data collection  (See Fig.~\ref{fig:data_collection}). Besides the benefits of allowing natural body motions and phone handling, this protocol exhibits two important changes to the nature of motion learning tasks.

\mybullet The positional ground-truth is obtained only for the 3D tracking phone attached to a harness, and our task is to estimate the trajectory of a body instead of the IMU phone. 

\mybullet The data offers a new task of body heading estimation.
A standard sensor fusion algorithm works well for the device orientation estimation~\cite{sabatini2006quaternion, madgwick2010efficient}. However, the body heading is more challenging, because the body orientation differs from the device orientation arbitrarily depending on how one carries a phone.
%
We collect the ground-truth body headings by assuming that they are identical to the headings of the tracking phone with an constant offset introduced by the misalignment of the harness.
We ask each subject to walk straight for the first five seconds, then estimate this offset angle as the difference between the average motion heading and the tracking phone's heading.

\mybulletend

We have made great engineering efforts in implementing the data processing pipeline to ensure high-quality sensor data and ground-truth, where we refer the details to the supplementary document. We conducted quantitative assessments of our system and found that our ``ground truth'' trajectories drift less than 0.3m after 10 minutes of activities. Similarly, the device orientation estimation from Android system API drift less than 20 degrees, while our system further reduces it to less than 10 degrees, which we treat as ground-truth during training.
Both IMU sensor data and 3D pose data are recorded at 200Hz. We also record measurements from the magnetometer and the barometer.


We divide the dataset into two groups: 85 subjects in group 1 and the remaining 15 subjects in group 2.
Group 1 is further divided into training, validating and testing subsets while group 2 is used to test the generalization capability of the model to unseen human subjects.

\section{Robust Neural Inertial Navigation (RoNIN)}

\begin{figure*}[tb]
    \centering
    \includegraphics[width=\linewidth]{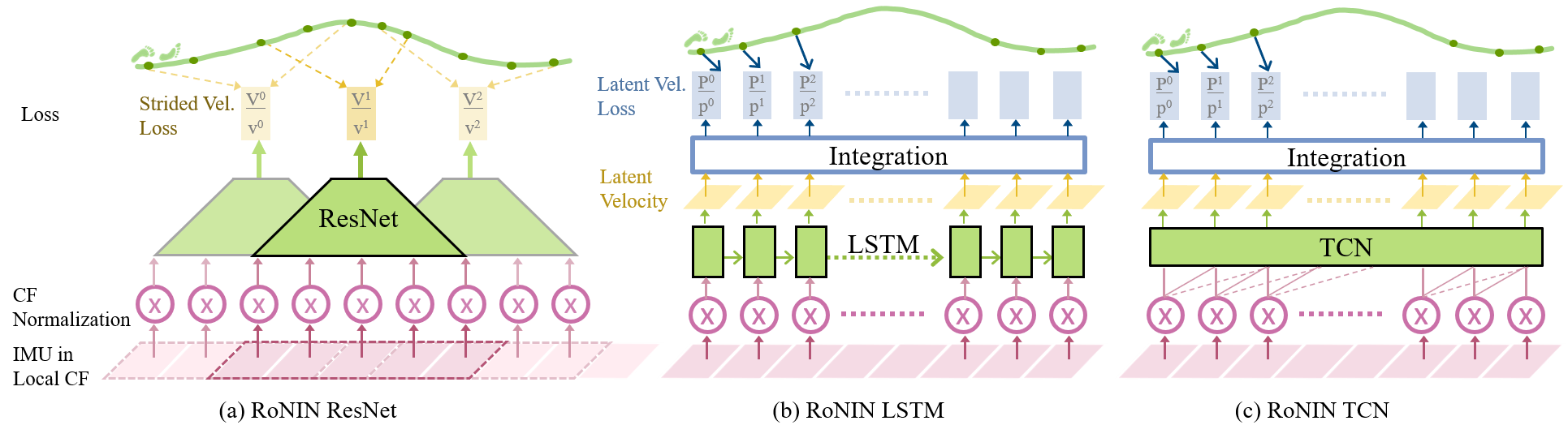}
    \caption{RoNIN Architectures: Neural network modules are in green and transformation layers are in white.}    \label{fig:networks}
\end{figure*}

Our neural architecture for inertial navigation, dubbed Robust Neural Inertial Navigation (RoNIN), takes
ResNet~\cite{resnet}, Long Short Term Memory Network (LSTM)~\cite{LSTM}, or Temporal Convolutional Network (TCN)~\cite{TCN} as its backbone. 
RoNIN seeks to regress a velocity vector given an IMU sensor history with two key design priciples:
1) Coordinate frame normalization defining the input and output feature space and 2) Robust velocity losses improving the signal-to-noise-ratio even with noisy regression targets. We now explain the coordinate frame normalization strategy, three backbone neural architectures, and the robust velocity losses. Lastly, the section presents our neural architecture for the body heading regression.


%

\subsection{Coordinate frame normalization}

Feature representations, in our case the choice of coordinate frames, have significant impacts on the training.
%
IMU sensor measurements come from moving device coordinate frames, while ground-truth motion trajectories come from a global coordinate frame.
RoNIN uses a {\it heading-agnostic coordinate frame} to represent both the input IMU and the output velocity data.



Suppose we pick the local device coordinate frame to encode our data.
The device coordinate changes every frame, resulting in inconsistent motion representation.
For example, target velocities would change depending on how one holds a phone even for exactly the same motions.

RIDI~\cite{yan2018ridi} proposed the {\it stabilized IMU coordinate frame}, which is obtained from the device coordinate frame by aligning its Y-axis with the negated gravity direction.
However, this alignment process has a singularity (ambiguity) when the Y-axis points towards the gravity (e.g., a phone is inside a leg pocket upside-down), making the regression task harder, usually completely fail due to the randomness.


RoNIN uses a heading-agnostic coordinate frame (HACF), that is, any coordinate frame whose Z axis is aligned with gravity. In other words, we can pick any such coordinate frame as long as we keep it consistent throughout the sequence.
%
%
The coordinate transformation into HACF does not suffer from   
singularities or discontinuities with proper rotation representation, e.g. with quaternion. 

During training, we use a random HACF at each step, which is defined by randomly rotating ground-truth trajectories on the horizontal plane. IMU data is transformed into the same HACF by the device orientation and the same horizontal rotation. The use of device orientations effectively incorporates sensor fusion~\footnote{We utilize Android's game rotation vector as device orientations.} into our data-driven system. At test time, we use the coordinate frame defined by system device orientations from Android or iOS, whose Z axis is aligned with gravity.

\subsection{Backbone architectures}

We present three RoNIN variants based on ResNet~\cite{resnet}, LSTM~\cite{LSTM} or TCN~\cite{TCN}.



\mysubsubsection{RoNIN ResNet}
We take the 1D version of the standard ResNet-18 architecture~\cite{resnet} and add one fully connected layer with 512 units at the end to regress a 2D vector.
At frame $i$, the network takes IMU data from frame $i-200$ to $i$ as a $200\times6$ tensor and produces a velocity vector at frame $i$. At test time, we make predictions every five frames and integrate them to estimate motion trajectories. 

\mysubsubsection{RoNIN LSTM}
We use a stacked unidirectional LSTM while enriching its input feature by concatenating the output of a bilinear layer~\cite{bilinear_layer}.
%
%
%
%
RoNIN-LSTM has three layers each with 100 units and regresses a 2D vector for each frame, to which we add an additional integration layer to calculate the loss. See 
Sect.~\ref{sec:loss_func} for the details of the integration layer.

\mysubsubsection{RoNIN TCN} TCN 
is a recently proposed CNN architecture, which
approximates many-to-many recurrent architectures with dilated causal convolutions.
RoNIN TCN has six residual blocks with 16, 32, 64, 128, 72, and 36 channels, respectively, where a convolutional kernel of size 3 leads to the receptive field of 253 frames. 
%

\subsection{Robust velocity loss}
\label{sec:loss_func}


Defining a velocity for each IMU frame amounts to computing the derivative of low-frequency VI-SLAM poses at much higher frame rate.
This makes the ground-truth velocity very noisy. We propose two  robust velocity losses that increase the signal-to-noise-ratio for better motion learning.

\mysubsubsection{Latent velocity loss}
RoNIN LSTM/TCN regresses a sequence of two dimensional vectors over time. We add an integration layer that sums up the vectors (over 400/253 frames for LSTM/TCN), then define a L2 norm against the ground-truth positional difference over the same frame-window. Note that this loss simply enforces that the sum of per-frame 2D vectors 
must match the position difference. To our surprise, both LSTM and TCN learn to regress a velocity in this latent layer before the integration, and hence, we name it the latent velocity loss.

\mysubsubsection{Strided velocity loss}
For RoNIN ResNet, the network learns to predict positional difference over a stride of 200 frames (i.e., one second), instead of instantaneous velocities. More specifically, we compute MSE loss between the $2D$ network output at frame $i$ and $P_{i} - P_{i-200}$, where $P_i$ is the ground truth position at frame $i$ in the global frame.




\subsection{RoNIN body heading network}


Different from the position regression, the task of heading regression becomes inherently ambiguous 
when a subject is stationary. 
Suppose one is sitting still at a chair for 30 seconds. We need to access the IMU sensor data 30 seconds back in time to estimate the body heading, as IMU data have almost zero information after the sitting event.
Therefore, we borrow the RoNIN LSTM architecture for the task, which 
is capable of keeping a long memory.

More precisely, we take the RoNIN LSTM architecture without the integration layer, 
and let the network  predict a $2D$ vector $(x, y)$, which are $\sin$ and $\cos$ values of the body heading angle at each frame. 
During training, we unroll the network over 1,000 steps for back-propagation. Note that the initial state is ambiguous if the subject is stationary, therefore we only update network parameters when the first frame of the unrolled sequence have velocity magnitude greater than $0.1 [m/s]$.

We use MSE loss against $\sin$ and $\cos$ values of ground-truth body heading angles. We also add a normalization loss as $\|1 - x^2 - y^2\|$ to guide the network to predict valid trigonometric values. 

\section{Evaluations: Preliminaries} \label{section:results}
We implement the proposed architectures using PyTorch and run our experiments using NVIDIA 1080Ti with 12GB GPU memory.


For RoNIN ResNet, we extract one training/validation sample (consisting of 200 frames) every 10 frames. Training samples are randomly shuffled for each epoch.
%
For RoNIN LSTM, we unroll the sequence to 400 steps once per $k$ frames, where $k$ is a random number between 50 and 150. Unrolled sequences are randomly batched to update network parameters. For RoNIN TCN, we construct one training/validation sample with 400 frames per $k$ frames, where $k$ is again a random number between 50 and 150.

For RoNIN ResNet (resp. RoNIN LSTM/TCN), we use a batch size of 128 (resp. 72), an initial learning rate of $0.0001$ (resp. $0.0003$), and ADAM optimizer while reducing the learning rate by a factor of 0.1 (resp. 0.75) if the validation loss does not decrease in 10 epochs, where the training typically converges after 100 (resp. 300/200) totalling 10 hours (resp. 40/30 hours). For linear layers we apply dropout with the keep probability 0.5 for RoNIN ResNet and 0.8 for RoNIN LSTM/TCN.





\subsection{Competing methods}
We conduct qualitative and quantitative evaluations of proposed algorithms on three datasets (RIDI, OXIOD, and RoNIN datasets) with four competing methods: 

\mysubsubsubsection{Naive double integration (NDI)} We transform linear accelerations (with gravity subtracted) into the global coordinate frame using device orientations and integrate them twice to get positions. We use Android API (Game Rotation Vector) to obtain the device orientations.

\mysubsubsubsection{Pedestrian Dead Reckoning (PDR)} We utilize a step-counting algorithm~\cite{tian2015enhanced} to detect foot-steps and move the position along the device heading direction by a predefined distance of $0.67m$ per step.

\mysubsubsubsection{Robust IMU Double Integration (RIDI)} We use the official implementation~\cite{yan2018ridi}. For RIDI and OXIOD datasets, we train a separate model for each phone placement type. For RoNIN dataset, where phone placement types are mixed, we train one unified model with $10\%$ of RoNIN training data, since their Support Vector Regression does not scale to larger dataset. Hyper-parameters are determined by a simple grid search on the validation set.



\mysubsubsubsection{IONet} We use our local implementation, as
the code is not publicly available.
As in RIDI method, we train a unified model on RoNIN dataset, and a separate model for each placement type for RIDI and OXIOD datasets.

\subsection{Device orientation handling}
PDR, RIDI and RoNIN rely on external device orientation information. For fairness we use the device orientation estimated from IMU for testing. During training, we use the same estimated orientations for RIDI dataset. For OXIOD dataset, we found that the estimated orientations are severely corrupted,~\footnote{We believe that this is due to the poor bias calibration and the inappropriate choice of APIs using the magnetic field, which is usually distorted in indoor.} and use the ground-truth orientations from Vicon during training. For RoNIN, we use the estimated device orientations if the end-sequence alignment error is below 20 degrees (See Fig.~\ref{fig:data_collection}), otherwise choose the ground-truth to minimize noise during training.




\subsection{Ground-truth alignment}
RoNIN estimates trajectories in the global frame and we directly compare against the ground-truth for evaluations. IONet is ambiguous in rotation and we use ICP to align the first 5 seconds of the estimated and ground-truth trajectories before evaluation in their favor. For NDI, PDR, and RIDI, we could align the estimated trajectory based on the device orientation at the first frame. However, a single frame information is often erroneous and we again use the first 5 seconds of the trajectories to align with the ground-truth.





\section{Evaluations}
We conduct comprehensive evaluations on two tasks: 1) position estimation among five competing methods on three datasets; and 2) body heading estimation by our method on the RoNIN dataset.

\begin{table*}[tb]
    \caption{Position evaluation. We compare five competing methods: Naive Double Integration(NDI), Pedestrian Dead Reckoning(PDR), RIDI, IONet, and RoNIN (3 variants) on three datasets: the RIDI dataset, the OXIOD dataset and our new dataset. The top three results are highlighted in red, green, and blue colors per row.
    }
    \centering
    \begin{tabular}{|c|c|c||r|r|r|r|r|r|r|}
        \hline
        \multirow{2}{*}{} & \multirow{2}{*}{Test subjects} & \multirow{2}{*}{Metric} & \multirow{2}{*}{NDI} & \multirow{2}{*}{PDR} & \multirow{2}{*}{RIDI} & \multirow{2}{*}{IONet} & \multicolumn{3}{c|}{RoNIN} \\ 
        \cline{8-10}
        &  & & & & & & ResNset & LSTM & TCN \\
        \hline \hline
        \multirow{4}{*}{RIDI Dataset} & \multirow{2}{*}{Seen} & ATE   & 31.06 & 3.52 & \color{blue}1.88 & 11.46 & \color{red}1.63 & 2.00 & \color{ForestGreen}1.66 \\
                                    \cline{3-10}
                                  &  & RTE   & 37.53 & 4.56 & \color{blue}2.38 & 14.22 & \color{red}1.91 & 2.64 & \color{ForestGreen}2.16 \\
                                    \cline{2-10}
                              & \multirow{2}{*}{Unseen} & ATE & 32.01 & 1.94 & \color{blue}1.71 & 12.50 & \color{ForestGreen}1.67 & 2.08 & \color{red}1.66 \\
                                    \cline{3-10}
                                    & & RTE   & 38.04 & \color{blue}1.81 & \color{ForestGreen}1.79 & 13.38 & \color{red}1.62 & 2.10 & 2.26 \\
        \hline \hline
        \multirow{4}{*}{OXIOD Dataset} & \multirow{2}{*}{Seen} & ATE  & 716.31 & \color{blue}2.12 & 4.12 & \color{red}1.79 & 2.40 & \color{ForestGreen}2.02 & 2.26 \\
                                    \cline{3-10}
                                    &  & RTE   & 606.75 & \color{blue}2.11 & 3.45 & \color{ForestGreen}1.97 & \color{red}1.77 & 2.33 & 2.63 \\
                                    \cline{2-10}
                                &    \multirow{2}{*}{Unseen} & ATE & 1941.41 & \color{ForestGreen}3.26 & \color{blue}4.50 & \color{red}2.63 & 6.71 & 7.12 & 7.76 \\
                                    \cline{3-10}
                                     & & RTE & 848.55 & \color{red}2.32 & \color{blue}2.70 & \color{ForestGreen}2.63 & 3.04 & 5.42 & 5.78 \\
        \hline \hline
        \multirow{4}{*}{RoNIN Dataset} & \multirow{2}{*}{Seen} & ATE   & 675.21 & 29.54 & 17.06 & 31.07 & \color{red}3.54 & \color{ForestGreen}4.18 & \color{blue}4.38 \\
                                    \cline{3-10}
                                &   & RTE   & 169.48 & 21.36 & 17.50 & 24.61 & \color{ForestGreen}2.67 & \color{red}2.63 & \color{blue}2.90 \\
                                \cline{2-10}
                                     & \multirow{2}{*}{Unseen} & ATE & 458.06 & 27.67 & 15.66 & 32.03 & \color{red}5.14 & \color{ForestGreen}5.32 & \color{blue}5.70 \\
                                    \cline{3-10}
                                &    & RTE & 117.06 & 23.17 & 18.91 & 26.93 & \color{blue}4.37 & \color{red}3.58 & \color{ForestGreen}4.07 \\
        \hline        
    \end{tabular}

    \label{tab:comparison}
\end{table*}

\subsection{Position evaluations}
We use two standard metrics proposed in~\cite{sturm2012benchmark}.

\mybullet Absolute Trajectory Error (ATE), defined as the Root Mean Squared Error (RMSE) between estimated and ground truth trajectories as a whole.

\mybullet Relative Trajectory Error (RTE), defined as the average RMSE over a fixed time interval, 1 minute in our evaluations.
For sequences shorter than 1 minute, we compute the positional error at the last frame and scale proportionally (e.g., double the number for a sequence of 0.5 minute).

\begin{figure*}[tb]
\centering
\includegraphics[width=\textwidth]{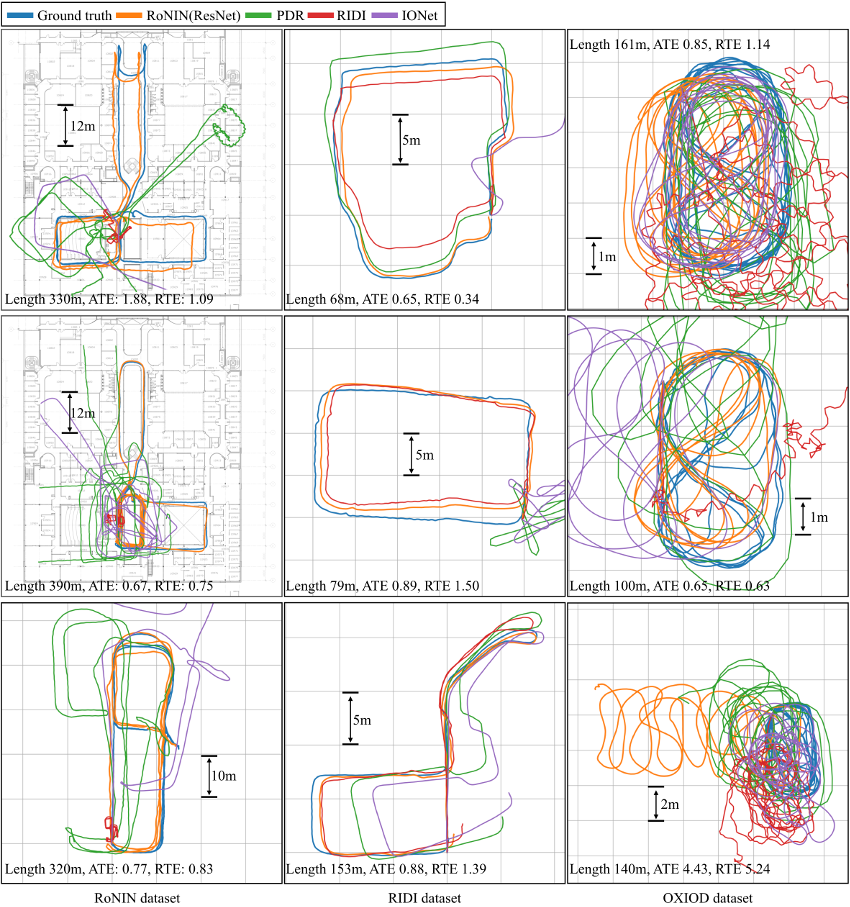}
\caption{Selected visualizations. We select 3 examples from each dataset and visualize reconstructed trajectories from all competing methods.
We choose RoNIN ResNet as our method, since its performance is more consistent across datasets. For each sequence we mark the trajectory length and report both ATE and RTE of our method. 
We also 
mark physical dimensions in all sequences to demonstrate that our method estimate trajectories with accurate scales. Three examples from our RoNIN dataset (left column) contains complex natural motions, where all other methods fail. Sequences from the RIDI dataset (middle column) contains hard motions. In particular, the middle example of the middle column contains extensive backward motion, where our method handles elegantly. Sequences from the OXIOD dataset (right column) are mostly short sequences with easy motions. However, our method gives large error for a few sequences (e.g. the bottom one) due to the large error in the provided device orientations.
}
\label{fig:traj}
\end{figure*}

\begin{figure*}[tb]
    \centering
    \includegraphics[width=\linewidth]{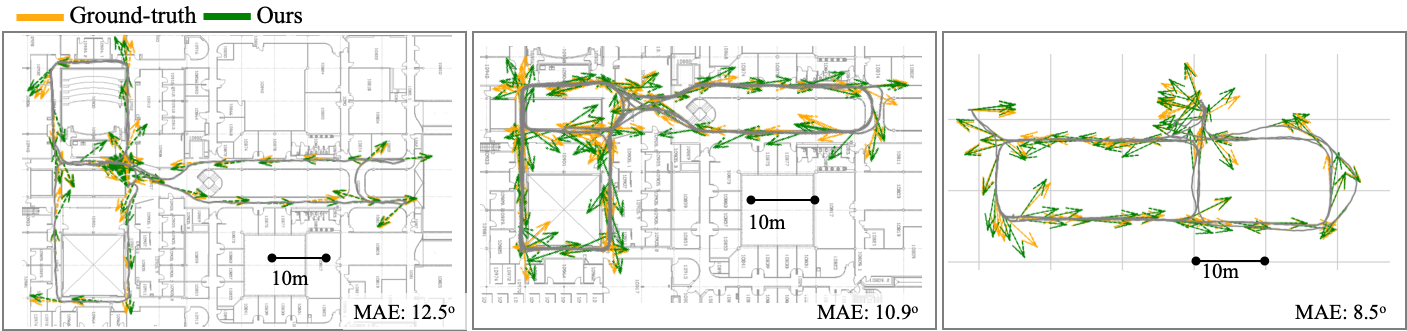}
    \caption{Selected visualization of heading angle estimations. 
    }
    \label{fig:heading_result}
\end{figure*}

\mybulletend
Table~\ref{tab:comparison} is our main result. All three datasets provide two testing sets, one for subjects that are also included in the training set and the other for unseen subjects. We report performance on both sets to evaluate the generalization ability. Figure~\ref{fig:traj} shows selected visualizations of the reconstructed trajectories against the ground-truth. We show RoNIN ResNet from our methods. Please refer to the supplementary material for more visualizations, including RoNIN LSTM/TCN. We exclude the visualization of NDI, whose trajectories are highly erroneous and would clutter the plots.
%
\begin{figure*}[tb]
    \centering
    \includegraphics[width=0.9\textwidth]{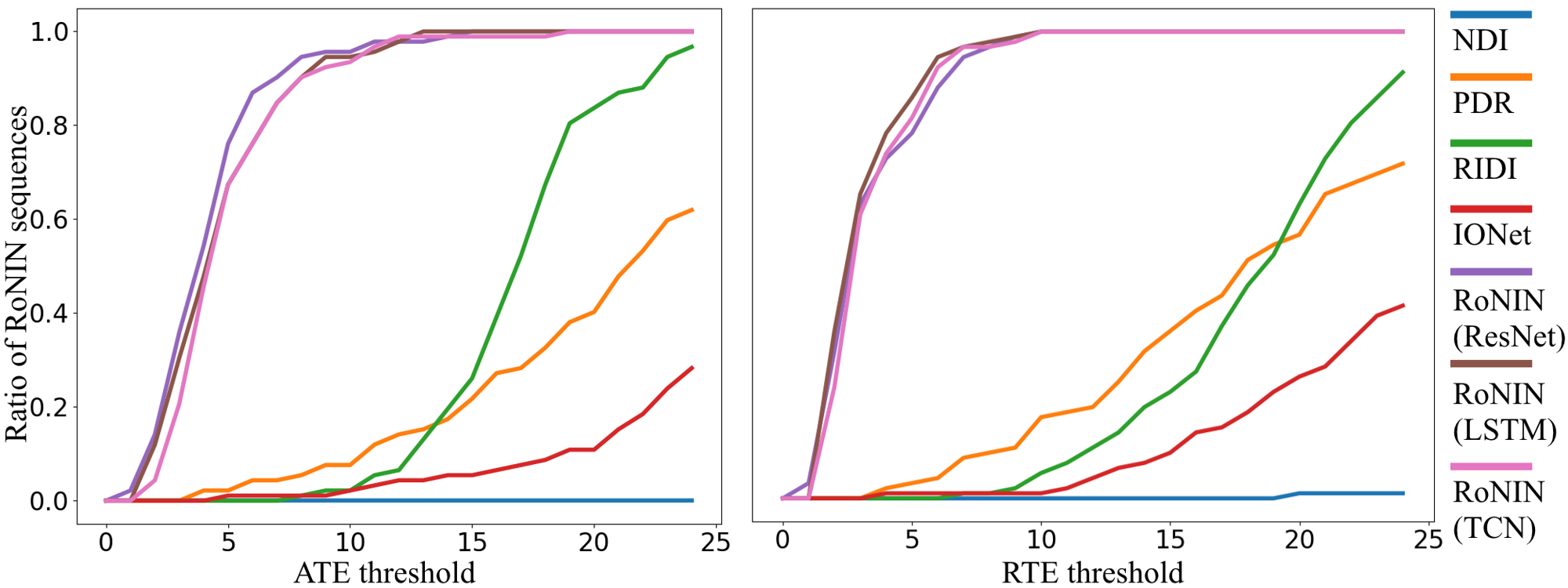}
    \caption{The ratio of RoNIN testing sequences under different thresholds on the two metrics. For instance, the left graph shows the ratio of sequences where the ATE is below a certain threshold 
    }
    \label{fig:curve}
\end{figure*}

Our method outperforms competing approaches on RIDI and RoNIN datasets with significant margins. Most notably, no previous methods can handle natural complex motions presented in our RoNIN dataset.
IONet suffers from accumulation errors in the motion heading estimation in the process of integrating the heading angle differences.
RoNIN LSTM and TCN perform slightly better than RoNIN ResNet on our dataset, but take 3 to 4 times more time to train.

Both RIDI and RoNIN struggle on the OXIOD dataset despite their easy motions, where even PDR works well. This is simply due to their poor device orientation estimations, which RIDI and RoNIN rely on and assume to be correct.
%
We expect their performance to improve with better bias calibration and the use of compass-free device orientation APIs.

Figure~\ref{fig:curve} shows the ratios of testing sequences in the RoNIN dataset (Y-axis) under different error thresholds on the two metrics (X-axis).
Our methods fail badly for a few sequences, where motions are not represented well in our training set (e.g. a phone in wildly moving handbag).

\subsection{Body heading evaluation}
We use two metrics for evaluations: 1) Mean Squared Error (MSE) of the $\sin/\cos$ representation of the heading; and 2) Mean Angle Error (MAE) of the estimated heading in degrees.
We compare against a simple baseline that reports the heading angles from the device orientations (i.e., device z-axis). We can evaluate only the heading difference in this baseline, and hence
align the device heading and the ground-truth body heading by first 5 seconds of the sequence.

Table~\ref{tbl.heading_result} and Figure~\ref{fig:heading_result} show the results. We notice that our errors become significantly larger (up to 20 degrees) for a few complex motion cases but are generally less than 12 degrees. The baseline fails because it does not account for the orientation difference between the device and the body.


\begin{table}[tb]
    \centering
    \caption{Body heading estimation task. We compare our RoNIN heading network with the baseline, which simply reports the device heading as the body heading.}
    \begin{tabular}{|l||r|r|r|r|}
    \hline 
                                   & \multicolumn{2}{l|}{Baseline} & \multicolumn{2}{l|}{RoNIN Heading} \\ \hline
        \multicolumn{1}{|l||}{Test Subjects} & Seen         & Unseen         & Seen       & Unseen \\ 
    \hline \hline
        \multicolumn{1}{|l||}{MSE}          & 1.58           & 0.99             & 0.06       & 0.08         \\ 
    \hline
        \multicolumn{1}{|l||}{MAE (degree)}          & 90.60    & 89.10       & 13.20       & 15.60  \\
    \hline
\end{tabular}
\label{tbl.heading_result}
\end{table}

\subsection{Ablation study} \label{sec:ablation}
\begin{figure}[tb]
    \centering
    \includegraphics[width=\columnwidth]{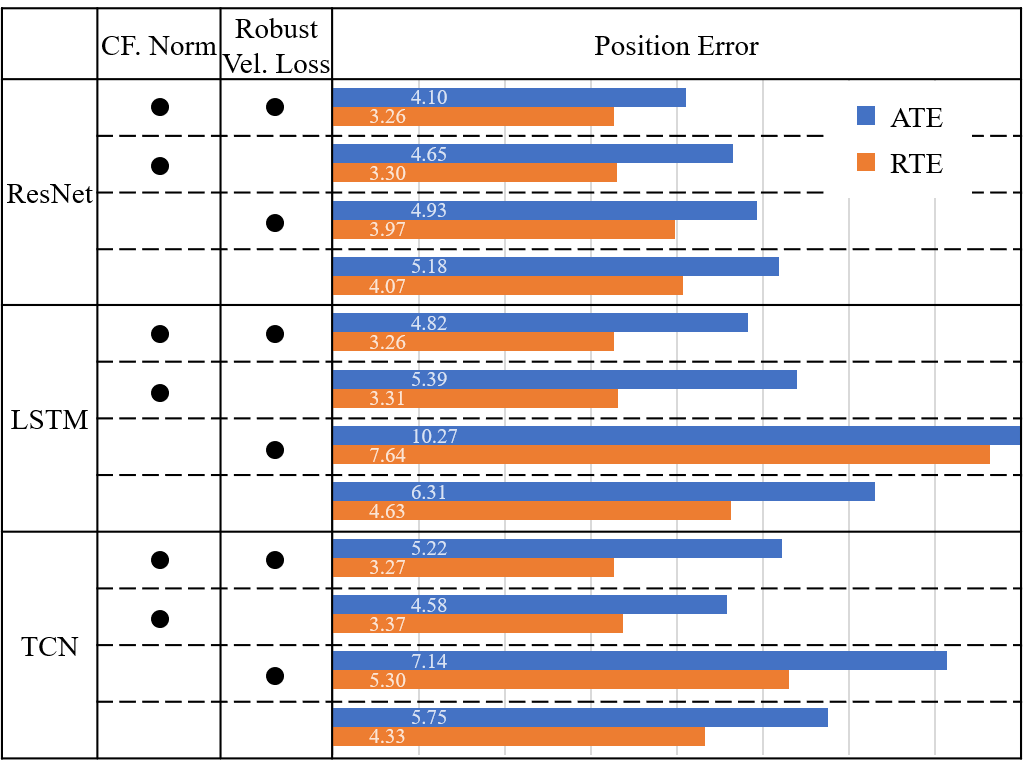}
    \caption{
    Ablation studies on coordinate frame normalization (CF. Norm) and the robust velocity losses (Robust Vel. Loss). Solid dots in each row means we enable the corresponding feature.
    }
    \label{fig:ablation}
\end{figure}

Figure~\ref{fig:ablation} shows the ablation study on the RoNIN dataset, demonstrating the effectiveness of the coordinate frame normalization and the robust velocity loss.
%
We evaluate the changes in the ATE and RTE metrics on the RoNIN test sequences (seen and unseen subjects combined) by turning on and off these two features for the three architectures.

\mybullet As a baseline without the coordinate frame normalization,
we supply the raw IMU sensor measurements as input and the ground-truth velocity in the local device coordinate frame as output.
%
Note that all three dimensions are needed and we rotate predicted velocities to the heading agnostic coordinate frame for position integration. The vertical axis is discarded during evaluations.

\mybullet
As a baseline without the strided velocity loss for RoNIN ResNet, we 
apply Gaussian smoothing with $\sigma=30$~\cite{yan2018ridi} to reduce the noise of ground-truth velocities, as suggested by RIDI. We use 
the smoothed instantaneous velocities as the supervision.
%

\mybullet As a baseline without the latent velocity loss for RoNIN LSTM/TCN,
we directly minimize MSE loss with the ground-truth instantaneous velocities.

\mybulletend
Figure~\ref{fig:ablation} demonstrates that the coordinate frame normalization and the robust velocity losses improve ATE and RTE overall, while the former seems to have larger impact. In particular, ATE and RTE shows the lowest errors when both features are combined except for one case, where ATE score of TCN is the second best with a very small margin.
\section{Discussions}
This paper sets a new foundation for data-driven inertial navigation research by 1) the new benchmark with large and diverse quantity of IMU-motion data as in real day-to-day activities; 2) new neural inertial navigation architectures making significant improvements over challenging motion cases; and 3) qualitative and quantitative evaluations of the current competing methods over the three inertial navigation datasets. 
The major limitation of our approach comes from the reliance on the device orientation estimations. The performance degrades significantly given data with poor device orientations, which is the main focus of our future work.
%
We will share all our code and data to promote further research towards an ultimate anytime anywhere navigation system for everyone's smartphone.



\section{Acknowledgement}
This research is partially supported by National Science Foundation under grant IIS 1618685, NSERC Discovery Grants, NSERC Discovery Grants Program Accelerator Supplements, and DND/NSERC Discovery Grant Supplement. We thank Ao Li for his contributions at an early stage of the project.

{\small
\bibliographystyle{ieee}
\bibliography{egbib}
}
\clearpage



\section*{Supplementary material}

# RoNIN: Robust Neural Inertial Navigation in the Wild: Benchmark, Evaluations, and New Methods
# Supplementary Material

Hang Yan*
Washington University in St. Louis
St. Louis, USA
yanhang@wustl.edu

Sachini Herath *
Simon Fraser University
BC, Canada
sherath@sfu.ca

Yasutaka Furukawa
Simon Fraser University
BC, Canada
furukawa@sfu.ca

The supplementary document provides the details of our data acquisition protocol: 1) Device preparation; 2) Data collection; and 3) Ground-truth preparation, and more qualitative and quantitative evaluations in Figs. 2, 3 and 4. Please also refer to our supplementary video which demonstrates a complex motion trajectory with the video-recording of the sequence with a variety of phone-handling motions.

## 1. Device preparation

We develop an in-house Android App for data collection, which can be installed on both the 3D tracking phone and the IMU phone. The App allows a pair of 3D tracking phone and IMU phone to be connected through Bluetooth for coarse time synchronization and centralized control.

We perform careful calibration of all sensors and provide calibration information, e.g. gyroscope biases, the accelerometer biases/scale sensitivities and magnetometer biaess, along with each data sequence. Our method uses calibrated raw IMU data instead of system provided linear accelerations and gyroscopes since we found that the latter exhibits differences across manufacturers.

We calibrate accelerometer's 3-axis biases and scale sensitivities with method presented in [1]. As a brief summary, we place the smartphone statically along different orientations. Calibration parameters are obtained by solving a non-linear optimization problem, in which the average magnitude of accelerations in each static period, when corrected, should be close to the gravity $g = 9.81m/s^2$. Each smartphone is re-calibrated once per day.

Gyroscope biases can be effectively estimated by placing the smartphone statically for 10 seconds, during which we average gyroscope readings as biases. We perform such calibration before and after each sequences.

We calibrate the magnetometer before each sequence by waving the smartphone along a 8-figure pattern and let the system to automatically update magnetometer biases. Magnetometer readings are still highly unreliable, even when calibrated, due to indoor magnetic disturbance. Therefore we disable it when obtaining device orientations. Leveraging noisy magnetometer data for better drifting reducing over a long period of time is one of our future interests.

## 2. Data collection

Each data sequence lasts between 4 and 10 minutes, during which testers can handle IMU phones freely under unconstrained motions. The only instruction for them is not to block the 3D tracking phone's camera, e.g. by the IMU phone, blank walls or other people. Time synchronization and spatial alignment are required for our two-device system to work properly.

**Time synchronization**: System clocks of two smartphones are coarsely synchronized when connected via Bluetooth. Before each sequence, we perform a horizontal rotation while rigidly attaching two phones together. We then estimate the precise time difference by signal correlation.

**Spatial alignment**: There are four coordinate frames involved in our two-device system: $L_T/L_I$ are local device frames of the 3D tracking phone and IMU phone, respectively. $W_T$ denotes the global frame of the 3D tracking phones, where ground-truth trajectories are defined and $W_I$ denotes the global frame of the IMU phone. Let $R_A^B$ denotes a rotation that rotate a 3D vector from frame $A$ to $B$. Our goal is to estimate rotation between two global frames, $R_{W_I}^{W_T}$, which is fixed for each sequence, so that IMU data and ground-truth trajectories can be represented in the same coordinate frame. Note that $R_{W_I}^{W_T}$ can be decomposed into Eq. 1:

$$R_{W_I}^{W_T} = R_{L_T}^{W_T} \cdot R_{L_I}^{L_T} \cdot (R_{L_I}^{W_I})^{-1} \qquad (1)$$

where $R_{L_T}^{W_T}$ is given by Tango's motion tracking and $R_{L_I}^{W_I}$ is given by Android's game rotation vector. We estimate $R_{L_I}^{T_I}$ by aligning two phones screen-to-screen, as shown in Fig. 1.

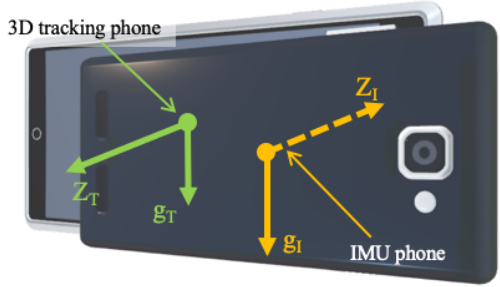

Figure 1. We attach two phones screen-to-screen for spatial alignment. The phone whose screen faces outside is the 3D tracking phone (green) and the one whose screen faces inside is the IMU phone (orange). $Z_T$ and $Z_I$ are local Z axes of two devices, which are exactly opposite to each other assuming screens are absolutely flat. $g_T$ and $g_I$ are gravity vectors in two local frames.

Assuming that screens are absolute flat and perpendicular to devices' Z axes, $R_{L_I}^{L_T}$ reduces to a 2D rotation around local Z axes, whose angle $\theta$ can be solved by Eq. 2:

$$\vec{g_T} = R_{L_I}^{L_T}(\theta) \cdot \vec{g_I} \quad (2)$$

where $\vec{g_T}$ and $\vec{g_I}$ are gravity vectors in $L_T$ and $L_I$.

At the beginning of each sequence, we attach two phones as shown in Fig. 1 while holding steady. $R_{W_I}^{W_T}$ is estimated for each frame within this period and then averaged.

## 3. Ground-truth preparation

Both ground-truth trajectories and IMU device's orientations suffer from drifting over time. We design procedures to reduce such drifting errors.

### 3.1. Refine ground-truth trajectories

Tango motion tracking fails in places where visual textures are poor, e.g. blank walls. We perform site survey and put checkerboard patterns at such places.

We leverage Tango's Area Learning functionality to reduce long-term drifting of ground-truth trajectories, where we first pre-scan the environment to create Area Description Files (ADF). Loop closure can be applied by relocalizing to ADF. Re-localization introduces discontinuities and may fail completely when subjects walks outside pre-scanned area. Therefore we record two trajectories with and without ADF re-localization. We manually select portions where re-localization fails and rely on incremental tracking in these regions. We evaluate ground-truth drifting by returning to a reference pose many times and have found that our refined trajectories drift less than 0.3m over 10 minutes, which is sufficient for our task.

### 3.2. Refine IMU orientations

We rely on Android's sensor fusion for IMU phone's orientations. We use an Android API (i.e., Game Rotation Vector), which does not utilize magnetometer. Gyroscope biases are estimated before each sequence. We further reduce rotational drifting by leveraging ground-truth information, using the filtering-based algorithm proposed in [2]. A standard kinematic model is used to predict 6DoF pose from raw accelerations and angular rates. Ground-truth positions from the 3D tracking phone are used as measurements to filter out optimal states, i.e. positions and orientations, and corresponding uncertainties.

IMU phone's orientations are not observable when the phone is in a pocket or a bag. We approximately evaluate the quality by endpoint rotational drifting, where we perform the same spatial alignment as described in Sec. 2 and compute angles between expected orientations and actual orientations, assuming Tango's orientations do not drift. We found that Android's sensor fusion typically drifts for 10 to 20 degrees over 10 minutes, while the filtering reduces it to 5 to 10 degrees. Note that filtered orientations are only used for training and we totally rely on Android's game rotation vectors during testing.

## References

[1] J. C. Lötters, J. Schipper, P. Veltink, W. Olthuis, and P. Bergveld. Procedure for in-use calibration of triaxial accelerometers in medical applications. *Sensors and Actuators A: Physical*, 68(1-3):221–228, 1998.

[2] A. Solin, S. Cortes, E. Rahtu, and J. Kannala. Inertial odometry on handheld smartphones. In *2018 21st International Conference on Information Fusion (FUSION)*, pages 1–5. IEEE, 2018.

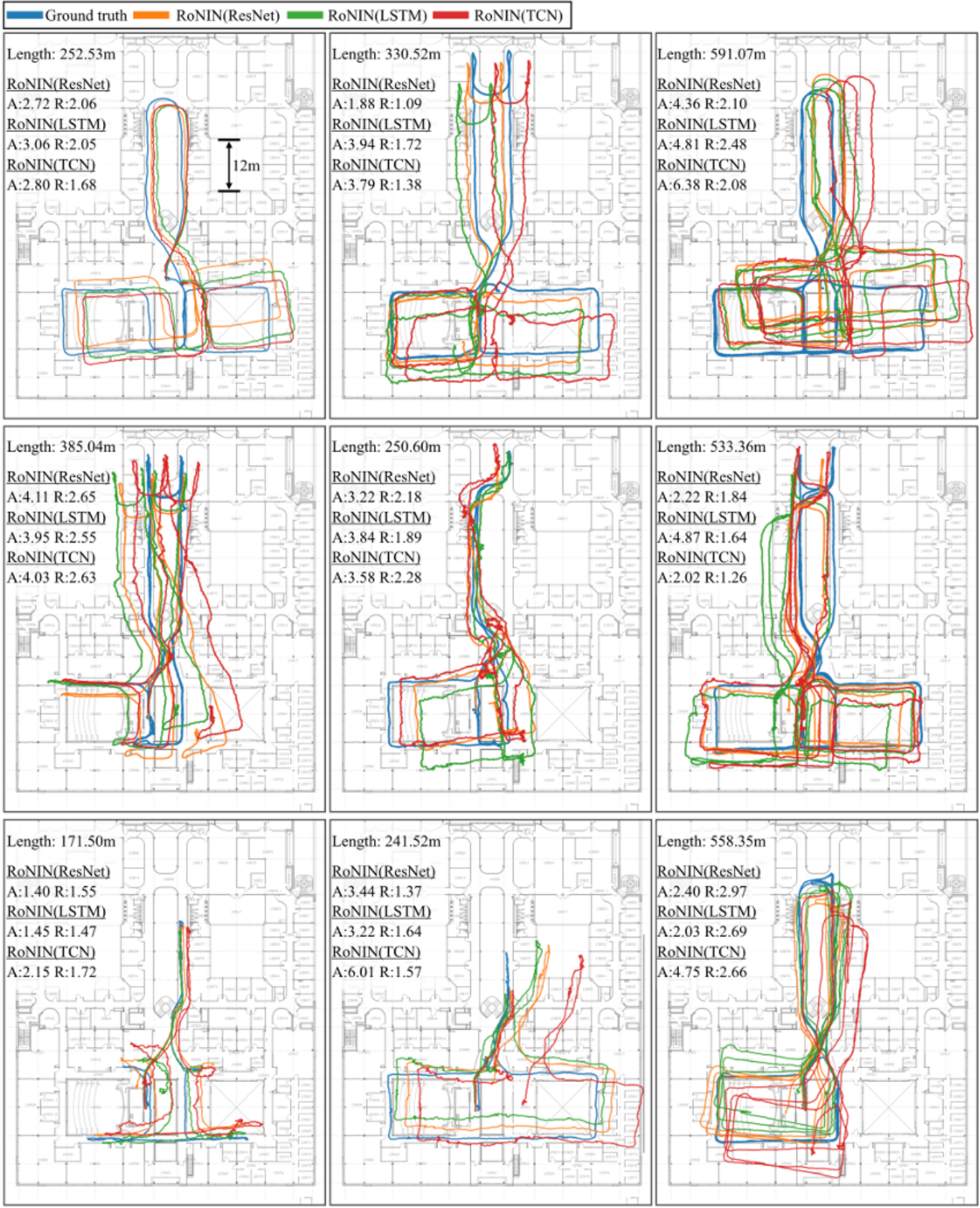

Figure 2. Selected visualizations of trajectories from 3 variants of RoNIN method on our dataset. Positional errors are marked within each figure, where "A" denotes ATE and "R" denotes RTE.

3

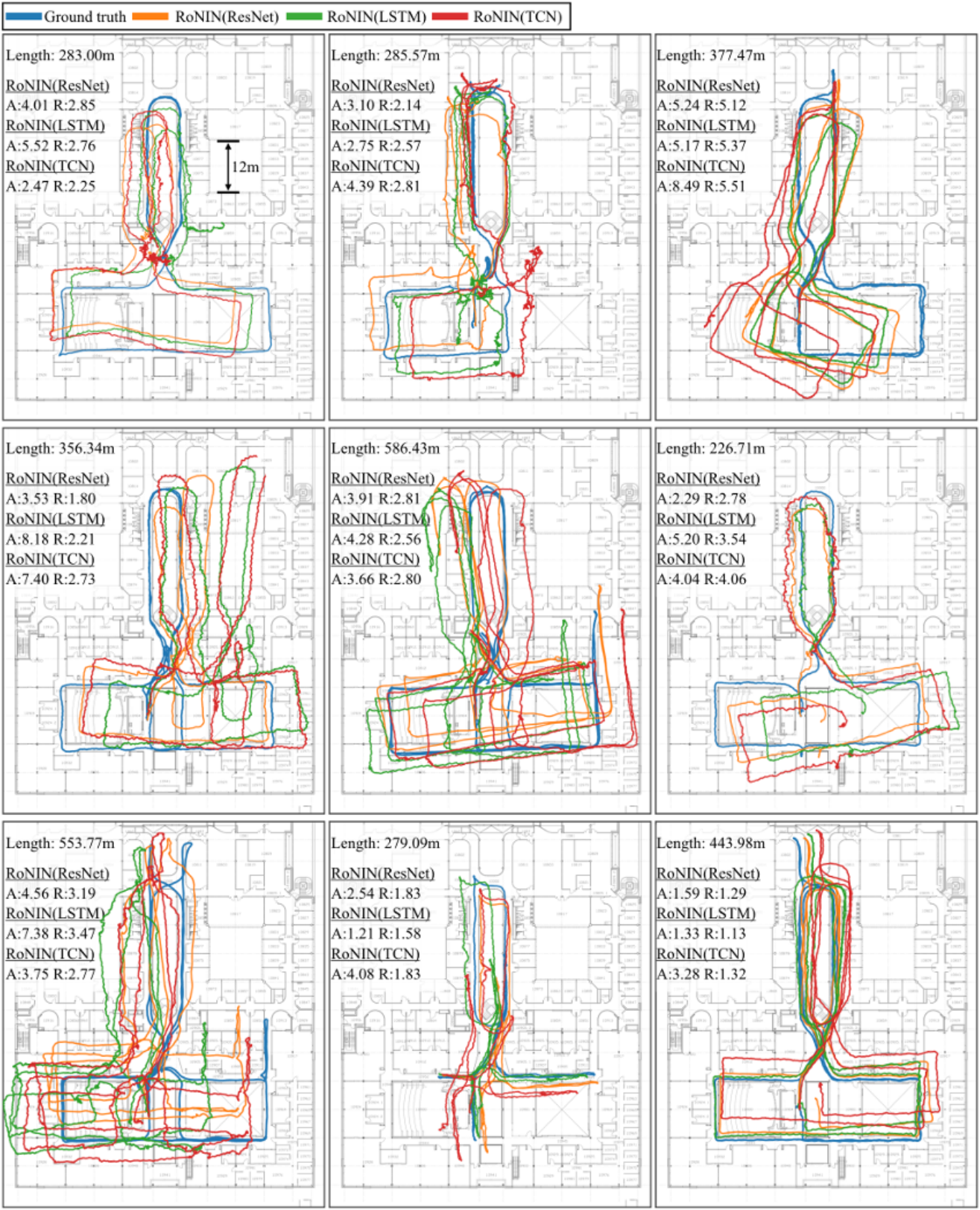

Figure 3. Selected visualizations of trajectories from 3 variants of RoNIN method on our dataset. Positional errors are marked within each figure, where "A" denotes ATE and "R" denotes RTE.

4

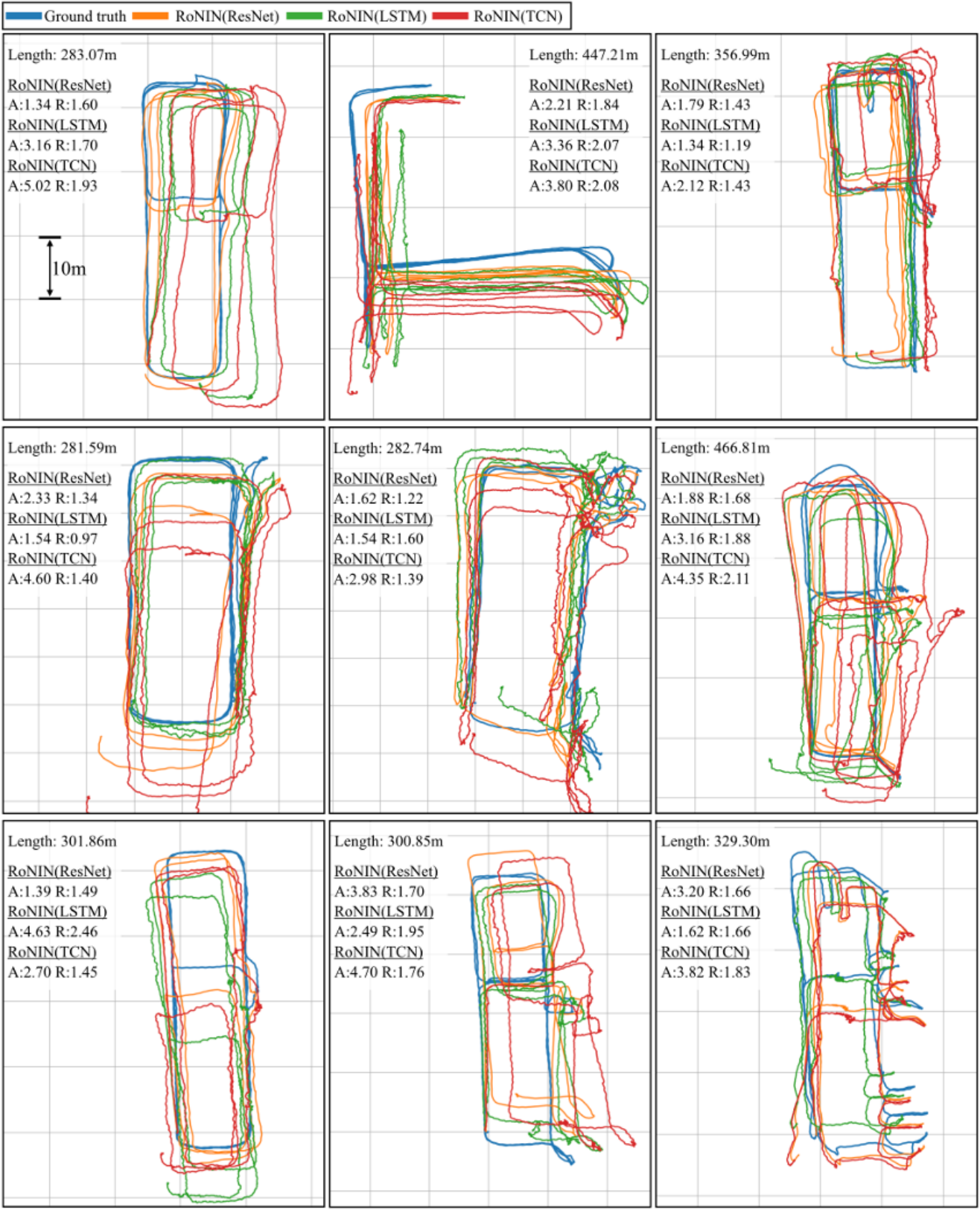

Figure 4. Selected visualizations of trajectories from 3 variants of RoNIN method on our dataset. Positional errors are marked within each figure, where "A" denotes ATE and "R" denotes RTE.

5


\end{document}